\documentclass[10pt,twocolumn,letterpaper]{article}

\usepackage{iccv}
\usepackage{times}
\usepackage{epsfig}
\usepackage{graphicx}
\usepackage{amsmath}
\usepackage{amssymb}
\usepackage{times}
\usepackage{epsfig}
\usepackage{graphicx}
\usepackage{amsmath, nccmath, mathtools}
\usepackage{amssymb}
\usepackage{paralist}
\usepackage[dvipsnames]{xcolor}

\usepackage{multirow, float, booktabs, boldline, hhline}
\usepackage{amssymb}
\usepackage{pifont}
\usepackage{physics}
\usepackage{bm}

\newcommand{\cmark}{\ding{51}}%
\newcommand{\xmark}{\ding{55}}%

\DeclareMathAlphabet\mathbfcal{OMS}{cmsy}{b}{n}

\newlength{\Oldarrayrulewidth} 
\newcommand{\Cline}[2]{%
	\noalign{\global\setlength{\Oldarrayrulewidth}{\arrayrulewidth}}%
	\noalign{\global\setlength{\arrayrulewidth}{#1}}\cline{#2}%
	\noalign{\global\setlength{\arrayrulewidth}{\Oldarrayrulewidth}}}

\def\eg{\emph{e.g}. } 
\def\ie{\emph{i.e}. }

\usepackage[pagebackref=true,breaklinks=true,letterpaper=true,colorlinks,bookmarks=false]{hyperref}

\iccvfinalcopy 

\def\iccvPaperID{6442} 

\ificcvfinal\fi

\begin{document}
	
	\title{TRiPOD: Human Trajectory and Pose Dynamics Forecasting in the Wild
	}
	
	\author{Vida Adeli$^1$, Mahsa Ehsanpour$^2$, Ian Reid$^2$, Juan Carlos Niebles$^3$, \\Silvio Savarese$^3$, Ehsan Adeli$^3$, Hamid Rezatofighi$^{4}$\\
		$^1$\textit{Ferdowsi University of Mashhad \quad $^2$University of Adelaide} \\ $^3$\textit{Stanford University \quad $^4$Monash University}\\ \normalsize{\url{https://somof.stanford.edu}}}
	

	\maketitle
	\ificcvfinal\thispagestyle{empty}\fi
	
	\begin{abstract}
		Joint forecasting of human trajectory and pose dynamics is a fundamental building block of various applications ranging from robotics and autonomous driving to surveillance systems. Predicting body dynamics requires capturing subtle information embedded in the humans' interactions with each other and with the objects present in the scene. In this paper, we propose a novel TRajectory and POse Dynamics (nicknamed TRiPOD) method based on graph attentional networks to model the human-human and human-object interactions both in the input space and the output space (decoded future output). The model is supplemented by a message passing interface over the graphs to fuse these different levels of interactions efficiently. Furthermore, to incorporate a real-world challenge, we propound to learn an indicator representing whether an estimated body joint is visible/invisible at each frame, \eg due to occlusion or being outside the sensor field of view. Finally, we introduce a new benchmark for this joint task based on two challenging datasets (PoseTrack and 3DPW) and propose evaluation metrics to measure the effectiveness of predictions in the global space, even when there are invisible cases of joints. Our evaluation shows that TRiPOD outperforms all prior work and state-of-the-art specifically designed for each of the trajectory and pose forecasting tasks. 
	\end{abstract}
	\section{Introduction}

The ability to forecast human movements (pose dynamics and trajectory) in time is an essential component for many real-world applications, including robotics \cite{mangalam2019disentangling,rosmann2017online}, healthcare \cite{kidzinski2020deep}, detection of perilous behavioral patterns in surveillance systems \cite{liu2018future,tang2020video}. 

\begin{figure}[!tbp]
  \centering
		\includegraphics[width=\linewidth]{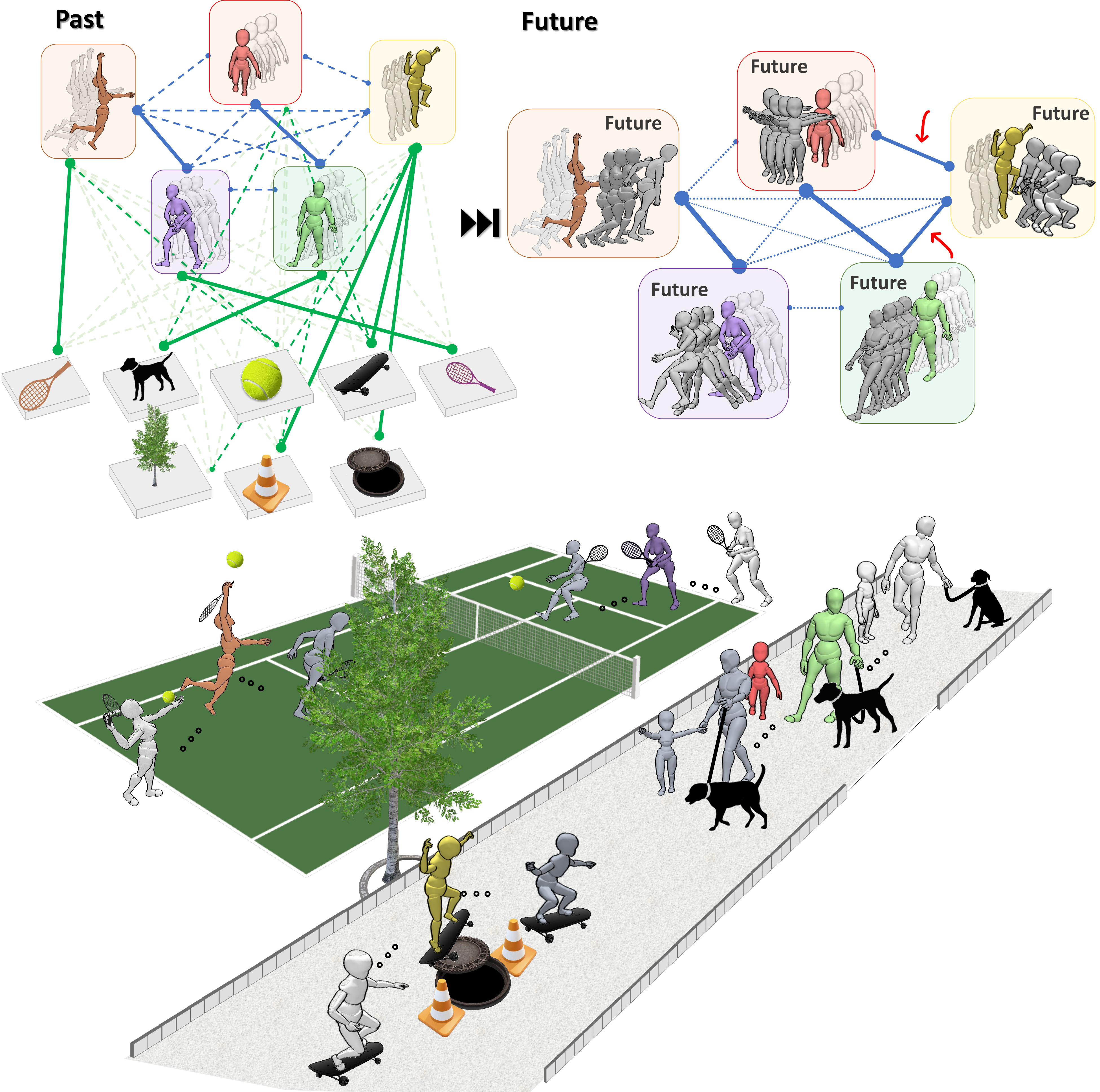}\vspace{-3pt}
	\caption{\footnotesize An example of a real-world scene containing different levels of interactions (human to human and human to objects). The top-left graph shows the weighted interaction graphs between humans (\textcolor{MidnightBlue}{blue} edges) and between humans and objects in the scene (\textcolor{ForestGreen}{green} edges). The top-right graph illustrates the evolved social interactions over time in the future. The \textcolor{red}{red} arrows indicate an example of a relation being intensified over time.}
	\label{fig:teaser}
	\vspace{-15pt}
\end{figure}  
While this problem sounds interesting, it is extremely challenging in real-world scenes due to the different factors involved. Humans are intuitively social agents, able to effortlessly conceive a detailed level of semantics from the scene, which contributes to making swift decisions for their next movements. To accurately forecast their trajectory and pose dynamics, one primary factor is the interactions between people in the scene and the influences their joints have on each other. For example, consider a tennis-playing scene, when the opponent starts serving and hits a stroke, the other person is probable to take a ready position in the near future (\eg see the purple agent in Fig.~\ref{fig:teaser}). Besides, the objects involved in the scene can provide informative clues for future prediction. For instance, when the person observes the ball in the tennis example, he/she would take a striking pose to return it. However, the movements of all the persons in the scene are not always highly correlated with each other nor the humans to objects. For instance, in Fig.~\ref{fig:teaser},  the pose and motion of tennis players will be barely affected by the skateboarder or his skateboard. This defines different levels of interactions that need to be discovered by the forecasting model. In addition, these different levels of interactions can change over time, \ie getting strengthen or weaken. In Fig.~\ref{fig:teaser}, lines thickened in future human-human graph (indicated by red arrows), show that the skateboarder's movements increasingly correlate with the parent and the kid (passengers), while some others lose correlations over time. 
Finally, a person might move outside the sensor field-of-view or be a partially/fully occluded by an object. In these cases, it is important to have an indication of visibility/invisibility for each prediction, which can be interpreted as its reliability score, conducive for the applications such as navigation safety and a collision risk assessment for an autonomous robot/vehicle. 

Existing solutions often neglect some of these challenging factors and hence fall short when applied to real-world \textit{in-the-wild} scenarios. 
Pose dynamics forecasting methods 
{\cite{chiu2019action, mo-att, Martinez17,wang2019imitation}} \textit{mostly} 
forecast the changes in joints with respect to a center position, ignoring the global position changes. They often do not effectively model \textit{all} the informative environmental and social interactions in the scene either. 
Similarly, the influence of individual joints is usually overlooked in trajectory forecasting 
{\cite{sgan, stgat}}. Moreover, existing frameworks often assume that all tracks and/or body joints are always observable in the past and future, which is an impractical assumption in many real-world scenarios.

To address these challenges, we push the current state of existing solutions for human pose dynamics and trajectory forecasting one step forward toward more practical scenarios in-the-wild by considering all these factors together. 
To this end, similar to other works that use attentional graphs for various purposes \cite{stgat,kosaraju2019social}, we model the input \textit{skeleton body joints}, the \textit{social human-human} and \textit{human-object interactions} with different attention graphs.
Since, these two types of information are different by nature, we give an effective solution to fuse them and as well make them insensitive to their choice of order by applying an \textit{iterative message passing}.
Furthermore, on account of the fact that humans may retain their influences on each other consistently in future, we do not content with only representing the history of interactions, but also we preserve their spatio-temporal attentional relationships by modeling them also in future prediction phase.
To overcome the problem of accumulative error in sequential models for long-term sequences and to speed up the convergence, we take a curriculum learning approach to train our model. {Finally, since there is no proper benchmark dataset for such real-world problem, we introduce a new \textit{benchmark} by repurposing existing datasets and introducing relevant evaluation \textit{metrics}.}

In summary, the main contributions of our paper are to 
\begin{inparaenum} [1)]
    \item   propose a model that considers all the mentioned challenges together by (\emph{i}) modeling the human skeleton, social and human-objects interactions through different dense and sparse graph incorporating attention,
    (\emph{ii}) introducing a \textbf{message passing} approach to efficiently fuse different level of interactions,
    (\emph{iii}) dynamically modelling the spatio-temporal attentional human interactions during \textbf{decoding phase},
    (\emph{iv}) addressing the concept of \textbf{joint invisibility or body disappearance} in trajectory and pose dynamics forecasting problem, (\emph{v}) suggesting a curriculum learning strategy to compensate accumulating error in recurrent models, \item  introduce \textbf{proper evaluation metrics} and a new \textbf{benchmark} for this  real-world problem. 
\end{inparaenum}

	\section{Related work}
\noindent\textbf{A.~Pose dynamics forecasting.}
Generally, pose dynamics forecasting aims to predict the future human pose coordinates in which global motion~(trajectory) is excluded. Early approaches modeled human dynamics by utilizing hand-crafted features and applying probabilistic graphical models~\cite{wang2008gaussian,wu2014leveraging}. Recently, deep sequence-to-sequence models~\cite{barsoum2017hp,chiu2019action,ghosh2017learning,jain2016structural,Martinez17,walker2017pose,wang2019imitation} have been used to capture such dynamics. Following the success of RNNs in capturing temporal dependencies, these models have been extensively used in capturing pose dynamics~\cite{chao2017forecasting,chiu2019action,fragkiadaki2015recurrent,Martinez17,rempe2020contact}. Since future forecasting of human pose dynamics is not a deterministic task, some works have utilized VAEs and GANs ~\cite{aliakbarian2020stochastic,Ho16,walker2017pose,yan2018mt,yuan2020dlow,zhou2018auto} and some focusing on the scene context \cite{cao2020long,corona2020context}. With the popularity of recently proposed transformers~\cite{vaswani2017attention, aksan2020spatio}, Mao et al.~\cite{mo-att} introduced an attention-based motion extraction model that aggregates current motion with its history. Likewise, in \cite{cai2020learning}, a transformer-based architecture is used for capturing the spatio-temporal correlations of the human pose.
All the aforementioned works are limited to only predicting local dynamics since global motion is subtracted from the human body joint coordinates. Further, interactions between joints in skeleton level and individual level are not modeled or captured. We argue that simultaneously capturing global and local dynamics is essential in forecasting reliable and robust 2D and 3D human poses and in general fine-grained human understanding. Moreover, human joints' movements are tightly coupled in the skeleton level and between interacting individuals. The problem is best formulated in a social manner. 

\noindent\textbf{B.~Human trajectory predictions.}
The goal of human trajectory prediction is to predict a set of 2D coordinates for each human characterizing its global motion. 
Human social interactions in crowds have always been considered an important cue for predicting humans' global trajectories, which were dominantly ignored by pose dynamic frameworks. Its literate goes back to pre-deep learning era when hand-crafted features were mainly used ~\cite{alahi2014socially,helbing1995social,mehran2009abnormal,pellegrini2009you,robicquet2016learning,yamaguchi2011you}. Although being successful, these works are task-dependent and require domain expert knowledge to carefully design hand-crafted rules. Recent deep data-driven models~\cite{alahi2016social,fernando2017soft+,gupta2018social,kosaraju2019social,lee2017desire,mangalam2020not, Sadeghian2018SoPhieAA, kothari2020human, amirian2019social} used a recurrent neural network and a social pooling layer on top to capture spatio-temporal social feature representation to predict the future trajectory of each individual.
{More recently, graph-structured models have been used to model human global motion and the existing interactions~\cite{corona2020context,huang2019stgat,ivanovic2019trajectron,kosaraju2019social,salzmann2020trajectron++, mohamed2020social, wang2021graphtcn}}.~\cite{huang2019stgat,kosaraju2019social} used graph attention networks~\cite{velivckovic2017graph} to model social interactions. Trajectron~\cite{ivanovic2019trajectron} proposed a graph-structured model that predicts many potential future trajectories. Trajectron++~\cite{salzmann2020trajectron++} also proposed a graph-based model that incorporates environmental information such as semantic maps and integrated with robotic planning. In \cite{giuliari2020transformer} a transformer-based method is proposed.
{Some other works improve the accuracy by incorporating novel trainable modules. For instance,~\cite{hu2020collaborative} proposed a neural motion message passing model to explicitly model the directed interactions between actors,~\cite{fang2020tpnet} proposed a trajectory proposal network to ensure safe and multimodal predictions and ~\cite{sun2020reciprocal} proposed reciprocal learning to train forward and backward networks.}  
While performing well, all these works lack modeling detailed human joints dynamics. Capturing and forecasting such fine-grained human body motions, \ie human poses, is essential for safe autonomous agents navigating through humans.

\noindent\textbf{C.~Pose dynamics and trajectory forecasting.}
As discussed earlier, learning fine-grained human joint dynamics as well as global motion are major components of a human understanding model that lead to development of a safe agent navigating in a crowd \cite{mangalam2019disentangling}. Recently, there have been some attempts to tackle the two problems in a unified manner. 
\cite{cao2020long} released a new synthetic dataset from a game engine 
and focused on utilizing scene context to tackle the unified task. \cite{adeli2020socially} unified human pose and trajectory forecasting in a socially-aware manner. 
Despite the novelty in the formulation of the problem by these works, social interactions are ignored or modelled in a basic way, making these works unable to handle invisible joints and complex in-the-wild scenarios. Here, we encode social interactions spatially and temporally via attention networks and message passing, and explicitly modeling human-objects interactions.   

	\section{Trajectory and Pose Dynamics Forecasting}
\begin{figure*}[!tbp]
\begin{center}
\includegraphics[width=\linewidth]{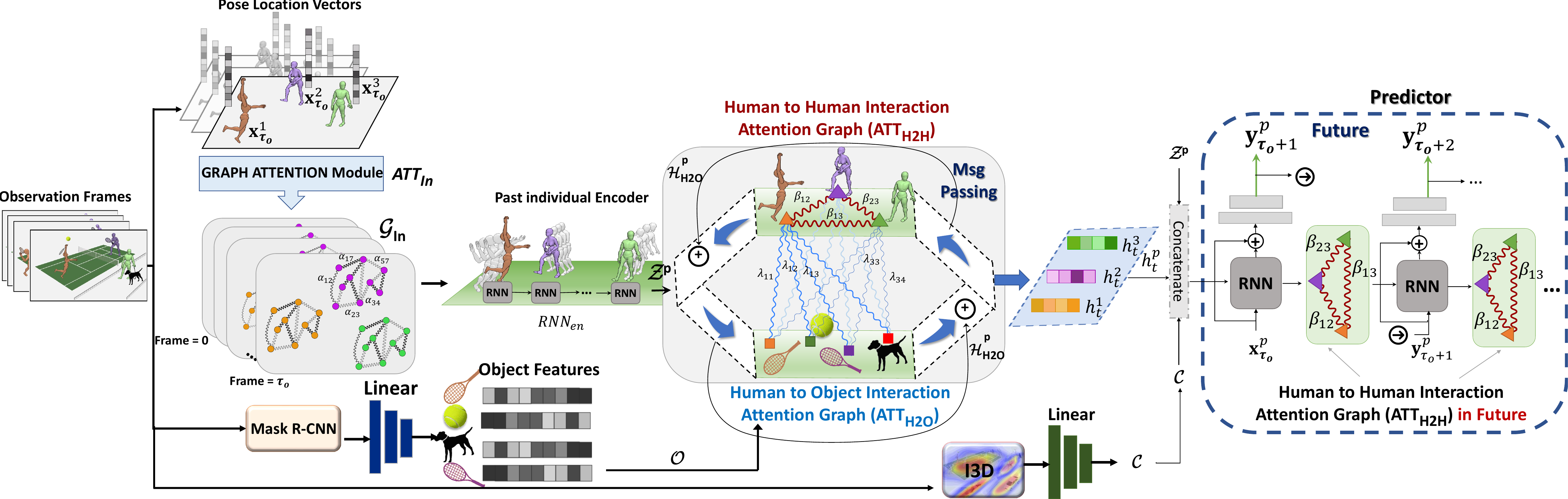}
\end{center}
\vspace{-10pt}
  \caption{\footnotesize An overview of the TRiPOD model proposed for human pose dynamics and trajectory forecasting in-the-wild. First, the history of poses are initially encoded using an attentive graph upon the body skeleton joints. Then, the encoded history is used to model interactions in Human to Human (H2H) and a Human to Object (H2O) attention graphs through a couple of iterative message passing. The future poses are then predicted, with the aid of the refined social interactions at every steps in the future.
  }
\label{fig:diag} \vspace{-12pt}
\end{figure*}

Humans are, by nature, social agents with complex interactions with not only other similar agents but also different parameters of the scene. All of these interactions and the conception of an agent from its surroundings build its actions, forming its future body pose and trajectory.
 Generally, the problem of joint human trajectory and pose dynamics forecasting can be defined as estimating the person's most probable future pose and trajectory given their prior history. Needless to say that when it comes to prediction in-the-wild, many other environmental factors come into play in addition to the individual's history.
Likewise, our goal is to model the complex human-human and human-object interactions in a way that can also predict  all the joint visibility indicators in the future.

\noindent\textbf{Problem Definition.}
Formally, assuming the past global history of a person $p \in \mathcal{P}$ as $\vb{X}_{1:\tau_o}^p = \{\mathbf{x}_1^p, \mathbf{x}_2^p, ..., \mathbf{x}_{\tau_o}^p\}$, where $\mathbf{x}_t^p \in \mathbb{R}^F$ with $F$ as the number of parameters describing the state of all the joints for person $p$, our goal is to predict the set of poses $\vb{Y}_{\text{\footnotesize +}1,\text{\footnotesize +}\tau_f}^p$ for the future $\tau_f$ frames. \useshortskip 
    \begin{equation} 
    \label{eq:output}
        \vb{Y}_{\text{\footnotesize +}1,\text{\footnotesize +}\tau_f}^p = \{\mathbf{y}_{\tau_o+1}^p, \mathbf{y}_{\tau_o+2}^p, ..., \mathbf{y}_{\tau_o+\tau_f}^p\}, \hspace{5 mm} \forall p \in \mathcal{P} 
    \vspace{-5pt}\end{equation} 
where $\mathbf{y}_t^p \in \mathbb{R}^F$. Throughout the paper, we consider any arbitrary variable, \eg $\phi_t$ being defined at time $t$ and $\phi_{t_1:t_2} = \{\phi_{t_1}, \phi_{t_1+1},..., \phi_{t_2}\}$ between times $t_1$ and $t_2$ and hold same convention for all variables. We also use the $\text{\small +}t$ notation as an indication of the future time. In our case, the state of each joint $k \in K$, is specified by 3 major indicators, \ie offset $\Delta \boldsymbol{\ell}$ (temporal location velocities), absolute locations $\boldsymbol{\ell}$ and joint visibility score $s$, which is a binary value being 0 if the joint is invisible, \ie $\mathbf{x}_t^p = \{(\Delta \boldsymbol{\ell}_{t}^p(k), \boldsymbol{\ell}_{t}^p(k),s_{t}^p(k)) \; |\; k\text{\small=}1:K\}$, 
where $\Delta \boldsymbol{\ell},\boldsymbol{\ell}\in\mathbb{R}^d ;\; s\in[0,1]$. $\mathbf{y}_t^p$ is defined the same as $\mathbf{x}_t^p$ (details are available in supplementary material). Unlike existing methods on pose forecasting, we use inputs in the original space, which means that the poses are not centered and contain the global trajectory. The importance of modeling these two sources is demonstrated in \cite{adeli2020socially}.
    

\noindent\textbf{TRiPOD Model:}
Our TRajectory and POse Dynamics (TRiPOD) model consists of multiple components (Fig. \ref{fig:diag}) and sub-components, 
described in detail as follows.

\noindent\textbf{A.~Attentional Human Pose History.}
     The dynamics of human body skeleton are the primary information that convey important knowledge for modeling the past history of pose and trajectory and also their prediction in future. This emphasizes the importance of how this information is represented to the pose forecasting models.

     Most earlier methods in pose forecasting \cite{chiu2019action,jain2016structural,martinez2017simple} utilize the joint coordinates, or other primary information upon coordinates such as velocities, to form a raw feature vector as their input. However, doing so ignores the significance of the natural connectivities in human body skeletons. 
     {Inspired by recent works \cite{li2020dynamic,zhao2019semantic}, we model skeleton pose as a graph, leveraging joint connections. However, as the influence of joints on each other is not uniform, we use an attentive graph encoder to model them.}
     The inputs to this pose attention graph $\mathit{ATT}_{In}(.)$ are state information of each body joint in each input frame $\vb{X}_{t}^p$. 
    \vspace{-5pt}\begin{equation}
     \label{eq:inputgraph}
        \mathcal{G}_{In, t}^p = \mathit{ATT}_{In}(\vb{X}_{t}^p; \vb{W}_{In}) 
    \vspace{-5pt}\end{equation} 
    where $\vb{W}_{In}$ is the set of the parameters of input pose attention graph. Accordingly, the output would be $\mathcal{G}_{In, t}^p$, which is a body pose representation, attending over different joint interactions. Then, an encoder $\mathit{RNN}_{en}$ (\eg LSTM) encodes the past history of each person's skeleton graph, up to time step $t$ as Eq. \eqref{eq:encoder}, with $\vb{W}_{en}$ as the encoder's parameters and $\bm{h}_{en, 0:\tau_o-1}$ as hidden states. 
    \vspace{-5pt}\begin{equation}
    \label{eq:encoder}
        \mathcal{Z}^p = \mathit{RNN}_{en}(\mathcal{G}_{In, 1:\tau_o}^p, \bm{h}_{en, 0:\tau_o-1}; \vb{W}_{en}) 
    \vspace{-5pt}\end{equation} 
    resulting in the encoded past global pose history $\mathcal{Z}^p$. 
    
\noindent\textbf{B.~Object and Global Scene Features.}
    As alluded earlier, to conceive the high-level semantics in the scene, the model should understand interactions between humans and objects and the scene context since the humans' pose is highly correlated to them. For this purpose, an object detector is used to extract the objects in the scene in the last observation frames, \ie $\tau_o$. 
    The final object representations ($\mathcal{O}$) is then obtained by passing the object feature vectors including visual feature, geometrical information, and its class label, through few embedding layers. Regarding the holistic scene features $(\mathcal{C})$ of all observation frames, we use a spatio-temporal model to represent the sequence and then the feature vectors are passed through a couple of embedding layers to form the final features. 

\noindent\textbf{C.~Human to Object Attention Module.}
    Since all the humans are not completely interrelated with all the objects in the scene, we also want the model to learn these different levels of interactions. To achieve this, we utilize a graph attention module ($\mathit{ATT}_{H2O}$) for encoding the human to object (H2O) interactions,  
    which takes as input the encoded past representation of person $\mathcal{Z}^p$ and the described object features $\mathcal{O}$ and outputs the H2O encoded interaction $\mathcal{H}^p_{H2O}$.
    \vspace{-4pt}\begin{equation} \label{eq:h2o}
        \mathcal{H}^p_{H2O} = \mathit{ATT}_{H2O}(\mathcal{Z}^p, \mathcal{Z}^{{P}\setminus p}, \mathcal{O}; \vb{W}_{H2O}) 
    \vspace{-4pt}\end{equation} 
    where $\mathcal{P}\setminus p$ represents all the people in the scene excluding person $p$ and $\vb{W}_{H2O}$ represents the parameters of H2O attention graph (\textcolor{MidnightBlue}{Blue} wavy links in Fig. \ref{fig:diag}).

\noindent\textbf{D.~Social Attention Module.} 
    Next, to augment the level of semantic in human pose dynamics forecasting in real-world, we encode humans' social interactions.
    {~\cite{huang2019stgat,kosaraju2019social} used graph attention networks~\cite{velivckovic2017graph} to model social interactions. However, they incorporated it with simplistic inputs, without considering object interactions.}
    Similar to the H2O attention module since different persons have different levels of interaction in the scene, their relation is modeled with an attention graph $\mathit{ATT}_{H2H}$, taking each person's representation produced by the $\mathit{ATT}_{H2O}$ graph and outputs a representation $\mathcal{H}^p_{H2H}$ 
    containing weighted social interactions. Similarly, $\vb{W}_{H2H}$ is the parameters of H2H attention graph (\textcolor{BrickRed}{Red} wavy links in Fig. \ref{fig:diag}).
    \vspace{-4pt}\begin{equation}
        \mathcal{H}^p_{H2H} = \mathit{ATT}_{H2H}(\mathcal{H}^p_{H2O}, \mathcal{H}^{\mathcal{P}\setminus p}_{H2O}; \vb{W}_{H2H}) 
    \vspace{-4pt}\end{equation} 
\noindent\textbf{E.~Message Passing.}
    So far, we have two different sources of information (human-human and human-object interaction) that are different by nature, even the types of effects they have are distinct.
    Thereby, an efficient approach is required to combine these two sources. We apply an iterative message passing inspired by approaches primarily proposed for obtaining useful information of molecular data  \cite{gilmer2017neural}. However, we reformed it to employ in our problem, to combine this two types of information effectively and make the framework invariant to their choice of order. 
    We extend the message passing concept to share information internally and between nodes of two different attention graphs.

    In simplistic terms, we describe our message passing module on the two undirected graphs of $\mathit{ATT}_{H2O}$ and $\mathit{ATT}_{H2H}$.
    Assume we have a set of node features $f$, where $f$ is the nodes in $\mathit{ATT}_{H2O}$ or $\mathit{ATT}_{H2H}$ graphs, and edge features of $e_{pu} \in \{\vb{W}_{H2H}, \vb{W}_{H2O}\}$, each node is then allowed to exchange information with its neighbours through a couple of $N$ time steps. Firstly, the node feature would become updated through a run of the H2O attention graph and is then fed to the H2H attention graph. We repeat this message passing process for a specified number of times $N$. After one such step, each node state would gain a primary perception of its immediate neighbors. Then, repeating more steps enhances these perceptions by incorporating second-order information and so on. Our message passing involves two main procedures: message passing (Eq. \eqref{eq:msg}) and the node update (Eq. \eqref{eq:msg_update}). During the message passing, each person's node's hidden state is updated based on the message $m^p_{n+1}$ while that of objects remains intact. 
    \vspace{-5pt}\begin{equation} \label{eq:msg}
        m^p_{n+1} = \underset{u\in neighbors(p)}{\mathit{ATT}}(f^p_n, f^u_n, e_{pu}),
    \vspace{-5pt}\end{equation}
    where $\scriptstyle  \mathit{ATT} \in \{\mathit{ATT}_{H2O}, \mathit{ATT}_{H2H}\}, \;\;
    p \in \{\mathcal{P}\}$ and $\scriptstyle  u \in \{\mathcal{O}, \mathcal{P}\}$. 
    \vspace{-5pt}\begin{equation} \label{eq:msg_update}
        f^p_{n+1} = \mathit{U}_n(f^p_n, m^p_{n+1}),
    \vspace{-5pt}\end{equation}
    $u$ denotes the neighbors of $p$ in H2O \& H2H graphs and $U_n$ is the update function at step $n$, here an average function.
    
    
\noindent\textbf{F.~Future Social Interactions.}
    To accurately predict future poses, only incorporating historical human social interactions is not sufficient.
    The model should also dynamically reconsider social interactions in the future, leveraging other people's actions during the same window of time. 
    {This valuable source of information, in addition to the loss supervision, can effectively improve training and performance during inference, as shown in the experimental section. This interactive decoding is mainly ignored by the previous works.}
    We address this issue by retaining the future interactions through the attentional H2H graph.
    Formally, after encoding all the previous individual and social history, scene and human-object interactions into a single representation for each person, the corresponding features are used as the input hidden state of a decoder predictor ($\bm{h}^p_{dec,0} = f^p_N$) to generate the set of future poses recursively after applying an embedding function $\psi$. 
     \vspace{-5pt}\begin{equation}
        y^p_{\text{\footnotesize +}t+1} = \psi\left(\mathit{RNN}_{dec}(y^p_{\text{\footnotesize +}t}, \bm{h}^p_{dec,\text{\footnotesize +}t}; \vb{W}_{dec}); \vb{W}_{\psi}\right) 
    \vspace{-5pt}\end{equation}
    Where $t \in (0,...,\tau_f-1)$ and
    $y^p_{\text{\footnotesize +}t+1}$ is the output global pose predicted for person $p$ at time $\tau_o+t+1$. Then the persons' representations are refined by the social attention graph forming the hidden state of the next time step (Eq. \eqref{eq:f-h2h(2)}) and the whole process continues until time step $\tau_f$. 
    \vspace{-3pt}\begin{equation}\label{eq:f-h2h(2)}
        \bm{h}^p_{+t+1} =  \mathit{ATT}_{H2H}(\bm{h}^p_{dec,\text{\footnotesize +}t}, \bm{h}^{\mathcal{P}\setminus p}_{dec,\text{\footnotesize +}t}; \vb{W}_{H2H})
    \vspace{-3pt}\end{equation} 

\noindent\textbf{G.~Training Strategies.}
A common problem in 
pose forecasting methods is that in the training phase, the model cannot recover from its accumulating errors at each time step and therefore, feeding this error as the input to the next step propagates it throughout the network and results in a large discrepancy between prediction and ground-truth poses in long-term. 
To address this problem, we \textit{first} make the final prediction to consider both the input and output of the RNN decoder at each time step using a skip connection to retain the output's continuity and can recover from the error. 
\textit{Second}, we employ the concept of curriculum learning \cite{bengio2009curriculum} and adopted it to train our model, which is starting with easier sub-tasks and gradually increasing the difficulty level of the tasks. This approach expedites the speed of convergence. Hence, we divide our future pose prediction problem for $\tau_f$ frames into $\frac{\tau_f}{\omega}$ sub problems where $\omega$ is the number of frames injected at each step. The model is first trained on the first sub-frames of prediction and after that, it learned this sub-task, then the second set of sub-frames are added to be trained. Note that the loss is calculated based on the new injected frames and the previous ones at each step.

\indent \textbf{Training Loss:}
    As is proven in \cite{adeli2020socially}, naturally the two problems of body pose dynamics forecasting and trajectory forecasting are highly correlated and should be approached jointly. Hence, we define a joint loss function in the global data coordinates and use the three described source of information as input (\ie offset $\Delta \boldsymbol{\ell}$, absolute locations $\boldsymbol{\ell}$ and invisibility indicator $s$). For the first two, we minimize the norm (the MSE $\ell_2$) error values of the ground-truth ($\Delta \boldsymbol{\ell}, \boldsymbol{\ell}$) and the prediction ($\hat{\Delta \boldsymbol{\ell}}, \hat{\boldsymbol{\ell}}$). 
    For the visibility score, we employ a Binary Cross Entropy loss. In training mode, if a joint is invisible in the truth, no gradient based on MSE loss ($\mathcal{L}_{\ell2}$) is calculated for it, while the visibility loss ($\mathcal{L}_{BCE}$) still penalizes the predictions. This concept is implemented by setting the value of loss to zero for that joints using a visibility mask $\mathcal{M}$ and a normalization factor $\eta$. Considering $\Theta$ as the collections of all weights, \ie $\Theta=\big(\vb{W}_{In},\vb{W}_{en},\vb{W}_{H2O},\vb{W}_{H2H}, \vb{W}_{dec}, \vb{W}_{\psi}\big)$, we use the following to train our model.
    
    \vspace{-5pt}\begin{equation}\label{eq:loss}
    \begin{split}
        \Theta^* = \underset{\Theta}{\text{argmin}} \; \mathbb{E}_{p,t} [\mathcal{L}(\vb{y}^p_t, \hat{\vb{y}}^p_t)]
    \end{split}
    \end{equation}
    \vspace{-15pt}
    \small \begin{equation}\label{eq:loss2}
    \begin{split}
        \mathcal{L}(\vb{y}^p_t, \hat{\vb{y}}^p_t) = \frac{1}{\eta}\Big(&\mathcal{L}_{\ell2}(\Delta \boldsymbol{\ell}_t^p,\hat{\Delta \boldsymbol{\ell}}_t^p) + \mathcal{L}_{\ell2}(\boldsymbol{\ell}_t^p,\hat{\boldsymbol{\ell}}_t^p) \Big) \times \mathcal{M} \\
        + & \mathcal{L}_{BCE}\left(s_t^p,\hat{s}_t^p\right)
    \end{split}
    \vspace{-5pt}\end{equation} \normalsize
    

\section{Benchmarking}
    There is no standard dataset available that can provide a fair pipeline for this problem, considering all the mentioned challenges. Furthermore, there is no metric for pose dynamics and trajectory forecasting, which accounts for joints' invisibility cases (such as occlusion or being outside the scene).
    We form a standard assessment platform as a benchmark, available at {\small\url{https://somof.stanford.edu}}, \emph{(i)} using existing multi-person datasets by repurposing them for human pose dynamics forecasting and \emph{(ii)} by proposing new metrics, taking the both source of errors, \ie predicted joint locations and visibility indicators, into account. 

\noindent \textbf{4.1. Metrics}
        Generally, consistent with prior work \cite{adeli2020socially,mo-att, Martinez17}, the fundamental procedure to report our evaluation results is based on the Mean Per Joint Position Error (MPJPE) \cite{ionescu2014human3}, which is the average Euclidean distance ($d_{\ell2}$) between ground-truth and estimated joint positions (but in our case in the global coordinate), averaged over number of persons in the sequence at each frame $i \in \{\tau_o+1, ..., \tau_o+\tau_f\}$. However, since we introduce the concept of invisible joints, we propose other types of metrics accounting for those cases.
        
        \noindent \textbf{Visibility-Ignored Metric (VIM)}. This metric is the simple MPJPE metric except that the invisible joints 
        (if exist) are not penalized and are simply discarded by considering truth.
        
\begin{figure}[!tbp]
          \centering
        		\includegraphics[width=4cm]{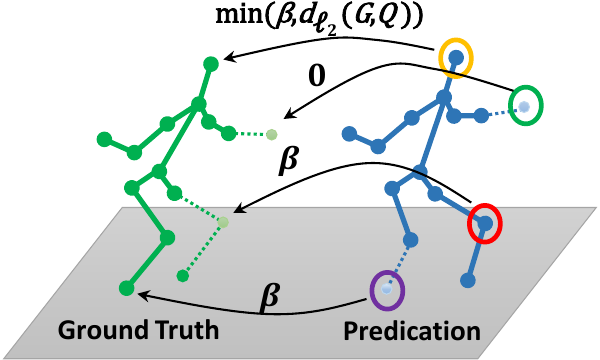}
        	\caption{\footnotesize Illustration of VAM (joint visibility-aware metric). Pale joints are the missing joints in ground-truth (\textcolor{ForestGreen}{green}) and prediction (\textcolor{MidnightBlue}{blue}) skeletons. The values on top of arrows are the penalty values. For other joints, the penalty is calculated by the minimum of error distance and cutoff $\beta$.}
        	\label{fig:metric}
        	\vspace{-10pt}
        \end{figure}

        \noindent \textbf{Visibility-Aware Metric (VAM)}.
        The second metric is proposed for performance evaluation in the presence of joint invisibility. Here, the goal is to calculate the distance of every joint per person in each time between the ground-truth and the prediction. When assuming the possibility of a joint being invisible, for every predicted joint, $q$, and its equivalent in ground-truth, $g$, three possible scenarios can be presumed (see Fig. \ref{fig:metric}):  1) Both joints are invisible. 2)  One joint is invisible in either the prediction or ground-truth. 3) Both joints are visible. These cases can be modeled by two singleton sets for each joint in ground-truth ($G=(\emptyset$ or $\{g\}$)) and prediction ($Q=(\emptyset$ or $\{q\}$)). To do so, inspired by the concept of miss-distance in OSPA metric \cite{schuhmacher2008consistent}, which is a distance metric (by mathematical definition) for comparing two sets of point patterns, we define $d^\beta_{o}(G,Q)$ as the distance between the two Singleton sets as follows:
        \vspace{-3pt}\begin{equation} \label{eq:VAM}
           d^\beta_{o}(G,Q) = \big( d^{(\beta)}(G,Q)^2+\beta^2|c_g-c_q|^2
           \big)^{\frac{1}{2}},
        \vspace{-2pt}\end{equation}
        where the $c_g\in\{0,1\}$ and $c_q\in\{0,1\}$ as 
        the cardinality of the two singleton sets, $d^{(\beta)}(G,Q) = \text{min}(\beta, d_{\ell2}(\{g\},\{q\}))$ is the distance between two joints (if both visible) cut off at $\beta>0$. Note $d^{(\beta)}(G,Q)$ is zero if any of the sets are empty.
        
        The value of $d^\beta_{o}(G,Q)$ for all three possible scenarios are shown in Fig. \ref{fig:metric}.
        Finally, the Visibility-Aware Metric (VAM) $d_v$ for all persons' $\mathit{K}$ joints is 
        \vspace{-5pt}\begin{equation} \label{eq:VAM2} \textstyle
           d_v = \frac{1}{\alpha}\sum_{p \in \mathcal{P}}\sum_{G,Q \in K}d^\beta_{o}(G^p,Q^p)
        \vspace{-5pt}\end{equation}
        where $\alpha$ is the normalization variable which can be defined as the summation of maximum cardinality of the prediction and ground-truth sets over each joint and each person: \;\; $\alpha = \sum_{p \in \mathcal{P}}\sum_{g,q \in K} max(c^p_g,c^p_q)$. \vspace{5pt}
        
        
    \noindent \textbf{Visibility Score Metric}. As the third metric, we evaluate the model only on the visibility scores $s$. For this purpose, we apply two criteria, the Intersection over Union (IoU) and the F1-score measure for all the joints in future frames, averages over the number of joints and  the number of persons.

\noindent \textbf{4.2. Data}
        Since the proposed method's main objective is to model and predict human poses in-the-wild, the choice of the dataset used for evaluation should also conform to the criteria in the real world. 
        To this end, 
        we re-purpose the recently released 3D Poses in the Wild dataset (3DPW)~\cite{3dpw} and the PoseTrack~\cite{posetrack} to create a standard pipeline for human pose dynamics and trajectory forecasting, to unify both communities, and to create a  platform with proper data splits and metrics to ensure a fair comparison between different approaches.
        These datasets reasonably provide us with the unconstrained set of information for complex real-world scenarios and contain both pose annotations and global trajectory data. The PoseTrack containing poses with invisible joints enables us to reconsider the occlusion and disappearing individuals problem in pose dynamics forecasting, which is essential to reflect the trustworthiness of the predictions. Details are elaborated in the supplement.

	\section{Experiments}
    
    In this section, we evaluate the performance of the TRiPOD model on the proposed benchmark 
    and compare it against state-of-the-art methods. We further conduct ablation study and provide some qualitative results.
    
    We use a one layer sequence-to-sequence model for encoding and decoding poses, with LSTM modules with a hidden dimension of 256. To model the attention mechanism in graph, we utilize the graph attention networks (GATs) \cite{velivckovic2017graph} which is dense in case of input pose and H2H module and sparse for H2O in which only humans and objects are linked. Also, the social graph in decoding phase has shared parameters with the H2H graph. 
    To extract the objects, a mask-rcnn-R-50-FPN-3x model, pre-trained on COCO with box AP of 41.0 
    \cite{wu2019detectron2} is used and for spatio-temporally representing the scene context, the I3D model \cite{I3D} pre-trained on Kinetics is employed. 
    Then, different two layer embedding modules are applied to the object and I3D features.
    The hyper-parameters are selected through experiments on a validation set 
    {(details are available in supplementary material)}. 
    To report the results, each experiment is performed three time and their average values are reported. 

    \subsection{Quantitative Results} \vspace{-5pt}
    \begin{table*}
    \scriptsize
    \begin{center}
    \caption{\footnotesize Error rate in \textbf{3DPW} (in cm) and \textbf{PoseTrack} (in pixel). In each column the best obtained result is highlighted with boldface typesetting.} \label{tlb:PoseTrack}
     \setlength{\tabcolsep}{3.5pt}
    \renewcommand{\arraystretch}{1.1}
    \begin{tabular}{c| c c c c c || c c c c c c c c c c c c}
    
        \Cline{0.8pt}{2-18}
        &\multicolumn{5}{c||}{\textbf{3DPW}}&\multicolumn{12}{c}{\textbf{\textcolor{black}{PoseTrack}}}  \\ 
        
        &\multicolumn{5}{c||}{\textbf{VIM (Invisibility ignored)}}&\multicolumn{5}{c}{\textbf{\textcolor{black}{VIM (Invisibility ignored)}}} &&&\multicolumn{5}{c}{\textbf{\textcolor{black}{VAM (Invisibility considered $\beta=200$)}}} \\ \Cline{0.8pt}{2-6} \Cline{0.8pt}{7-11} \Cline{0.8pt}{14-18}
 
        \multirow{2}{*}{}&\multicolumn{5}{c||}{prediction time in milliseconds}&\multicolumn{5}{c}{prediction time in milliseconds }&&&\multicolumn{5}{c}{prediction time in milliseconds}\\
        &100&240&500&640&900&80&160&320&400&560&&&80&160&320&400&560\\
        \hlineB{2}
        
        Center pose \hspace{0.3cm} Trajectory &&&&&&&&&&&&&&  \\ 
        
        PF-RNN~\cite{Martinez17} + S-LSTM~\cite{alahi2016social}  &73.2&126.79& 180.03&201.75& 277.53&87.05&103.35&129.19&138.66&160.96&&&101.37&114.09&133.88&145.39&160.49\\ 
        PF-RNN~\cite{Martinez17} + S-GAN~\cite{sgan} & 68.40 & 119.72 & 172.73 & 195.88 & 263.05&84.74&98.94&121.35&129.55&150.16&&&99.96&111.51&129.3&140.08&154.09 \\
        PF-RNN~\cite{Martinez17} + ST-GAT~\cite{stgat} &67.12 & 116.53 & 164.61 & 189.82 & 250.88&80.93&95.72&119.03&127.66&149.44&&&96.16&109.06&127.5&137.75&152.49 \\
        \hline
        
        Mo-Att~\cite{mo-att} + S-LSTM~\cite{alahi2016social} &65.24&109.67&168.94&200.16&268.14&84.45&101.63&121.16&135.48&157.48&&&100.02&113.89&130.41&144.27&158.24 \\
        Mo-Att~\cite{mo-att} + S-GAN~\cite{sgan}  &63.41&106.25&161.89&193.98&258.51&81.33&97.45&118.74&125.78&147.12&&&99.21&109.56&129.08&139.25&152.47  \\
        Mo-Att~\cite{mo-att} + ST-GAT~\cite{stgat}  &62.41&94.59&153.24&188.02&249.91&78.14&93.75&115.61&119.31&140.83&&&97.16&107.42&125.36&136.04&149.78 \\
        \hline\hline
        
        \multicolumn{1}{c|}{Joint Trajectory \& Pose} &&&&&&&&&&&&&&  \\
      \multicolumn{1}{c|}{SC-MPF \cite{adeli2020socially}} & 45.44&73.73&129.23&159.47&208.31&21.41&39.92&66.32&77.73&93.41&&&82.74&95.67&106.5&113.19&133.78 \\
        \hline\hline
        
        \multicolumn{1}{c|}{\textcolor{black}{\textbf{TRiPOD}}} & \textbf{31.04}&\textbf{50.8}&\textbf{84.74}&\textbf{104.05}&\textbf{\text150.41}&\textbf{15.36}&\textbf{26.32}&\textbf{46.45}&\textbf{57.94}&\textbf{71.78}&&&\textbf{50.62}&\textbf{60.77}&\textbf{79.69}&\textbf{80.07}&\textbf{96.98} \\
        \hline
    
    \end{tabular}
    \end{center} \vspace{-20pt}
    \end{table*}
     
    \noindent \textbf{A.~Baselines.}
    Generally, the two problems of pose dynamics and trajectory forecasting had been commonly treated as isolated problems by the community and thereby the number of approaches that jointly model these information are limited. Consequently, to investigate the effectiveness of the proposed method, we break down the problem and retrain the task with the models in each community separately and then combine their results in prediction. Various works have been conducted in these two community, however, we try to select the most popular and recent state-of-the-art methods that are conceptually similar to our problem and could be simply applied. Ultimately, we select \cite{mo-att, Martinez17} and \cite{alahi2016social, sgan, stgat} as the most popular and recent state-of-the-art methods in human pose dynamics and trajectory forecasting, respectively, and did our best to fairly retrain these methods by effectively setting up their parameters and prepare data in compliance to how should be used to obtain the best results (data preparation details for baselines are available in supplementary material). We also compare against SC-MPF \cite{adeli2020socially} that considers the two problems jointly. Note, other prior works do not mainly consider the joint problem.  
    
    \begin{figure}[!tbp]
          \centering
        		\includegraphics[width=5.5cm]{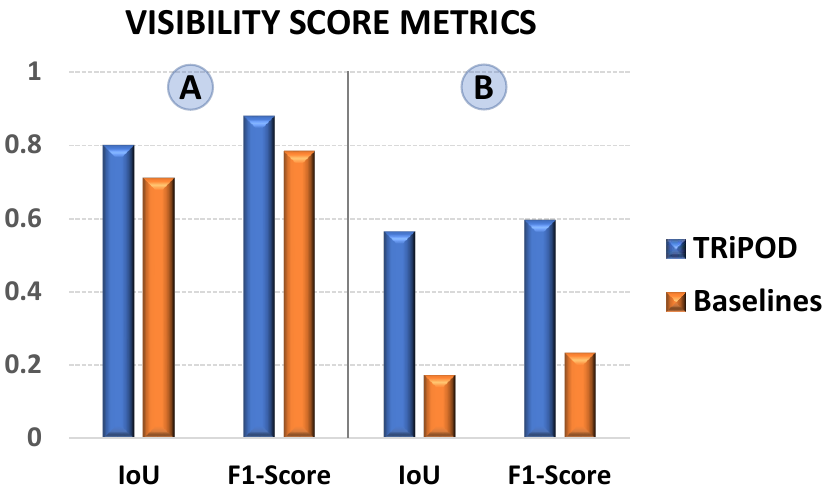}
        	\caption{\footnotesize Comparison of visibility score metrics. (A): All data considered. (B): Joints with at least one case of disappearance in future are considered.}
        	\label{fig:vsm}
        	\vspace{-15pt}
        \end{figure} 
    \noindent\textbf{Joint Evaluation:} We first compare the results jointly in global space and in the next step represent the comparison in each problem separately. For \textit{3DPW}, the results are reported based on visibility ignore metric (VIM), since it does not have invisible joint cases (in this case, VIM acts the same as simple MPJPE metric). 
    Table \ref{tlb:PoseTrack} evidences that we achieve the best results for joint pose dynamics and trajectory forecasting on 3DPW. For \textit{PoseTrack}, since the dataset contains cases in which the joints are occluded or poses disappear by the persons leaving the scene, we employ all the three proposed metrics (see subsection 4.1) for its evaluation.
    Table~\ref{tlb:PoseTrack} shows that we consistently outperform other methods in both ignored and considered joint visibility metrics (VIM and VAM) in PoseTrack. Fig.~\ref{fig:vsm} demonstrates the evaluation results for the visibility score metric. The visibility scores $s$ for baselines are considered to be always true, since they assume all joints are visible during the whole past and future.
    The goal is to investigate the performance of the model in recognizing visible/invisible cases. To do so, The \textit{IoU} and \textit{F1-score} of the binary vectors ($s$) are calculated (as described in subsection 4.1). Then two approaches are adopted: \textbf{A)} The whole data is used in the metric evaluation. Generally, in PoseTrack, $27.28\%$ and $28.82\%$ of joints are invisible in the observation and the future frames, respectively.
     \textbf{B)} Since the visible cases are more frequent, to better show the gap between the performance of TRiPOD and the baselines in predicting invisibility, we perform evaluations only on joints with at least one future case of invisibility. This experiment shows the performance difference between methods when some joints disappear in some future frames.
    Fig.~\ref{fig:vsm} shows that TRiPOD is able to estimate joint invisibility, and this claim can be better seen when always-present joints are not considered in the evaluation.

    
    \begin{table}
    \scriptsize
    \begin{center}
    \caption{\footnotesize Ablation study on \textbf{3DPW} based on VIM (ignored invisibility). Each notation is defined as: $\mathbfcal{C}$: Scene context, $\mathbfcal{P}$: Input pose representation, tensors (T) or attention graph (G). $\mathbfcal{H}$: Social module (max operation (M) or attention graph (G)). $\mathbfcal{O}$: Human-object graph. $\mathbfcal{M}$: Message passing. $\mathbfcal{FH}$: Human interactions in future. \textbf{CL}: Curriculum Learning. \vspace{-5pt}}   \label{tlb:ablation} 
     \setlength{\tabcolsep}{2.5pt}
    \renewcommand{\arraystretch}{1.1}
    \begin{tabular}{c|c c c c c c||c c c c c}
    
        \hlineB{2}

        \multirow{2}{*}{}&\multirow{2}{*}{$\mathcal{C}$}&\multirow{2}{*}{$\mathcal{P}$}&\multirow{2}{*}{$\mathcal{H}$}&\multirow{2}{*}{$\mathcal{O}$}&\multirow{2}{*}{$\mathcal{M}$}&\multirow{2}{*}{$\mathcal{FH}$}& \multicolumn{5}{c}{prediction time in milliseconds }\\
        
        &&&&&&&100&240&500&640&900\\
       
        \hlineB{1.5}
        S-MPF ~\cite{adeli2020socially} & \xmark & T & M & \xmark & \xmark &\xmark  &52.89&89.27&146.2&176.98&249.18 \\
        SC-MPF ~\cite{adeli2020socially} & \cmark & T & M & \xmark & \xmark &\xmark  &45.44&73.73&129.23&159.47&208.31 \\
        \hline\hline
        
        \textcolor{black}{Baseline 1} & \cmark & T & G & \xmark & \xmark &\xmark  &39.74&64.44&106.13&128.36&181.32 \\
        
        \textcolor{black}{Baseline 2} & \cmark & G & G & \xmark & \xmark &\xmark  &33.99&54.57&93.75&114.75&167.32 \\

        \textcolor{black}{Baseline 3} & \cmark & G & G & \cmark & \xmark &\xmark  &\textcolor{black}{32.64}&\textcolor{black}{52.73}&\textcolor{black}{91.25}&\textcolor{black}{111.9}&\textcolor{black}{166.68} \\

        \textcolor{black}{Baseline 4} & \cmark & G & G & \cmark & \cmark &\xmark &\textcolor{black}{32.85}&\textcolor{black}{52.64}&\textcolor{black}{88.77}&\textcolor{black}{108.38}&\textcolor{black}{161.72}  \\
        \hline\hline
        \textcolor{black}{TRiPOD} & \cmark & G & G & \cmark & \cmark &\cmark &\textcolor{black}{\textbf{31.56}}&\textcolor{black}{\textbf{51.97}}&\textcolor{black}{\textbf{86.53}}&\textcolor{black}{\textbf{107.52}}&\textcolor{black}{\textbf{153.12}}  \\
        
        \textcolor{black}{TRiPOD(CL)} & \cmark & G & G & \cmark & \cmark &\cmark &\textcolor{black}{\textbf{31.04}}&\textcolor{black}{\textbf{50.8}}&\textcolor{black}{\textbf{84.74}}&\textcolor{black}{\textbf{104.05}}&\textcolor{black}{\textbf{150.41}}  \\
        
        
        \hlineB{2}

    \end{tabular}
    \end{center} \vspace{-20pt}
    \end{table}
        
\begin{figure*}[!tbp]
    \centering
	\includegraphics[width=16cm]{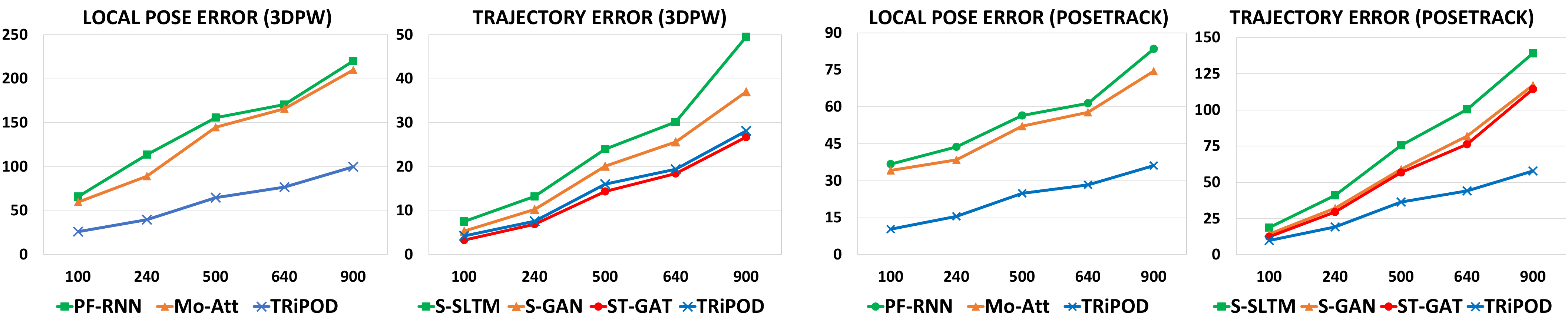}
	\caption{\footnotesize The VIM error rate 
	in each pose dynamic and trajectory forecasting problem separately, for 3DPW and PoseTrack in different times. For the disjoint problems the VIM becomes same as the FDE metric for trajectory information.}
	\label{fig:chart}
   \end{figure*}

    \noindent\textbf{Separate Evaluation:} For more comparison, we also evaluate the TRiPOD model on center pose and trajectory independently and compare results with baselines in each community. 
    Fig.~\ref{fig:chart} illustrates 
    that our TRiPOD model achieves the lowest error rate (VIM) for center pose prediction in both datasets and in trajectory forecasting performs in par with ST-GAT in 3DPW and outperforms others in PoseTrack (a more challenging dataset with invisible joints).
    
    \noindent\textbf{Results Discussion:} 
    The results in Table \ref{tlb:PoseTrack} and Fig.~\ref{fig:chart} reveal that although using the combination of two state-of-the-art methods in each community (Mo-Att+ST-GAT) can improve results, the outputs for the naive joint learning method SC-MPF proves that the tasks of pose and trajectory forecasting are interrelated and results can be significantly improved when they are modeled jointly. Finally, the TRiPOD shows its superiority by jointly modeling the two tasks and incorporating effectively different levels of historical and futures interactions in the scene and allowing the model to be aware of the possibility of joint invisibility.
   
   \begin{figure*}[!tbp]
  \centering
		\includegraphics[width=16cm]{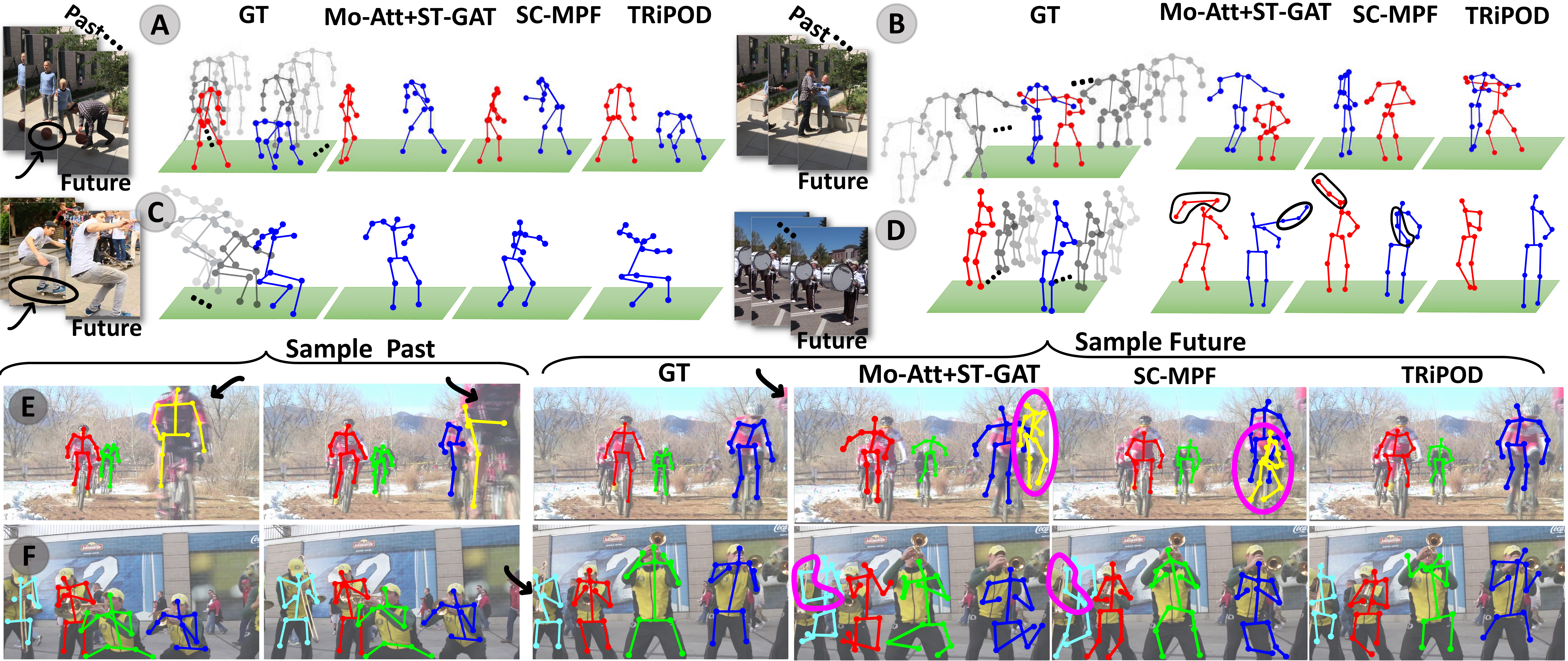}
	\caption{\footnotesize Qualitative results of TRiPOD, SC-MPF and Mo-Att+ST-GAT. (A) and (B) are samples taken from 3DPW and the others are from PoseTrack.}
	\label{fig:qualitative}
\end{figure*}  
   
    \noindent \textbf{B.~Ablation Study.}
    In particular, we examine each component's contributions in TRiPOD by performing an ablation study on the 3DPW in Table \ref{tlb:ablation}.
    The first two rows are the results for the SC-MPF baseline that uses a max-pooling social operation, and the second set of results (baselines 1 to 4) are for experiments in which each component is added to the model one by one, showing the effect of each module in the final performance. The results indicate that every module improves the prediction results which evidences the benefits of exploiting different levels of semantic from the scene both in observation and prediction. We also performed an ablation study on the number of iterations for message passing and the results reported in this table are based on three iterations (the results for this ablation are available in the supplementary material). Finally, the best performance is obtained by training using a curriculum learning scheme.

    \subsection{Qualitative Results}
    To better understand the contribution of the TRiPOD model in improving the understanding of different interactions in the scene, occlusion, or termination of pose existence, we visualize the prediction results for a number of samples, comparing the TRiPOD against SC-MPF and Mo-Att+ST-GAT outputs. Fig.~\ref{fig:qualitative} (\textit{A}) and (\textit{C}) illustrate cases that evidence the effect of interpreting the interactions between humans and objects in the scene. Being aware of the objects and the person's history, the model could effectively predict the final pose in the future very close to that of the truth compared to the baselines' prediction. Similarly, case (\textit{B}) shows the same effect when the interactions between humans are modeled effectively in TRiPOD. Sample (\textit{D}) further evidences the importance of the model being aware to estimate occlusion. Instead of outputting improper outlier predictions (joints in black curves in (\textit{D})), the model is capable of recognizing occlusion cases with the help of both occlusion handling indicators and also the interpretation of the object and its location in the scene. Finally, the two bottom cases (\textit{E} and \textit{F}) show an agent leaving the scene or the joints being out of the camera sight. TRiPOD is favorably capable of handling such cases. The pink curves indicate such faulty predictions in sample future prediction.

	\vspace{-5pt} 
	\section{Conclusion}
In this paper, we proposed a model for joint human pose dynamics and trajectory forecasting. Instead of only paying attention to the individual's history, our model considers different levels of semantics and interactions in the scene by attentionally modeling skeleton pose, social and human-object interactions through different graphs, and incorporating global context. The model also reinforces the future predictions, letting them be socially inter-correlated in the future in each time-step. Our method is also able to handle occlusion and pose disappearance cases. The accumulative error problem in long-term sequences is effectively handled through training model in a curriculum concept. Finally, we introduce a benchmark and relevant metrics to jointly solve the pose dynamics and trajectory forecasting problem in more realistic scenarios. Our experiments demonstrated that our TRiPOD model outperforms state-of-the-art methods in this problem. Directions for future works can be incorporating 3D information (when camera parameters are available) and considering multi-modal future predictions.

\noindent\textbf{Acknowledgement.} J.C. Niebles and E. Adeli would like to thank Panasonic for their support.

	{\small
		\bibliographystyle{ieee_fullname}
		\bibliography{refs}
	}

	\clearpage

	\section*{A. Discussion}
	\noindent\textbf{Why two separated H2H and H2O graphs?}
	In this section, we discuss the reasons for considering the human to human (H2H) and human to objects (H2O) as two different graphs. \textit{First}, these two sources of information are naturally different and the type of information and influences obtained from them are also disparate. Therefore, considering them as similar nodes of a single graph is not intuitively a sensible practice. \textit{Second}, densely connecting these two different types of information as a single huge graph and training them all together makes it difficult for the model to converge, increases the model's complexity and the overall computation. Besides, the quality of the final features obtained are not necessarily effective. Therefore, a better practice is to consider the H2H and H2O as two different graphs but devising a solution to effectively fuse these two sources of information and their effects (described as iterative message passing in the paper).

	\section*{B. Benchmark Data Details}
	Here, we provide more details about the two datasets that we used and re-purposed to create our human pose dynamics and trajectory forecasting benchmark.
	
	\noindent\textbf{3D Poses in the Wild (3DPW)} \cite{3dpw}: The recently released 3DPW is a challenging outdoor dataset captured using IMU sensors, with a moving camera and consists of 60 long video clips divided into 3 train, test and validation splits. We divided the video clips into multiple non-overlapping 30-frame shorter sequences sampling over every two frames resulting in 342 sequences and to investigate the importance of predicting pose dynamics and trajectories in complex scenarios, we only consider the multi-person sequences containing social interactions. We use the 3 provided splits, However, switched the train and test splits since the number of sequences in test have become larger after the aforementioned preprocess. The body poses are in world coordinate and the results are reported in centimeter (cm). In 3DPW, the pose annotations are represented by 3D locations of 24 body joints. Since some of the joints, such as fingers and toes, are not important for the current problem, we limit our selection to a subset of 13 main body joints including the neck, shoulders, elbows, wrists, knees, hips, and ankles.  In 3DPW, we feed 1000ms of past history into the model and the goal is to predict the next 1000ms of future data.
	
	\noindent\textbf{PoseTrack} \cite{posetrack}: The PoseTrack is a large-scale multi-person dataset which covers a diverse variety of interactions including person-person and person-object in dynamic crowded scenarios. In PoseTrack, pose annotations are provided for 30 consecutive frames centered in the middle of the sequence. The pose forecasting in this dataset is challenging because of the wide variety of human actions in real-world scenarios and the large number of individuals in each sequences with large body motions and a high number of occlusions and disappearing individuals cases. Since this dataset contains cases with huge portion of joints being invisible during the time, we perform some preprocess steps to make it practicable for the current problem. We maintained only those persons that are not completely invisible in all the observation frames (means at least some partial past history should be available for a person to enable the model forecasts its future). Moreover, there were some faulty, inaccurate annotations in the dataset that we did our best to refine them. The overall number of sequences is 516 which are from the training split of this dataset. We use 60\% of these sequences for training our model and the rest were split equally for validation and test. We use a set of 14 joints in 2D space defining the poses including the head, neck, shoulders, elbows, wrists, knees, hips, and ankles. The data being used is in image coordinate and therefore the results are reported in pixel. In PoseTrack sequences, we trained our model by observing the past 560ms frames and learning to minimize the prediction error over the next 560ms.
	
	\section*{C. Input Data Types}
	As mentioned in the paper, we used both the offset and absolute positions as the model's input data. We practically investigated that using both offset and absolute provides the best results. The reason is that although the offset is zero mean and improves the training process, a small error in offset prediction can deviate significantly from the absolute value in high dimensions or in a long time horizon. On the other hand, the absolute is not zero mean value but keeps offset error bounded to the absolute position. Considering both information together can recompense the mutual errors.
	
	\section*{D. Baselines Setups}
	The Posetrack containing invisible joints entails some initial setups for the baselines (center pose \cite{Martinez17, mo-att} or trajectory forecasting\cite{alahi2016social, sgan, stgat}) to make it possible for them to be trained on this dataset. For training the baselines with both datasets, pose information is first centered by subtracting the neck position from every joint and the pose dynamics forecasting methods \cite{Martinez17, mo-att} are trained on the local poses of the datasets. Simultaneously, the  trajectory, considered as neck positions, is also learned by the three state-of-the-art trajectory forecasting methods \cite{alahi2016social, sgan, stgat}. Then, during prediction, we add the trajectory predictions to the local pose to obtain the global poses (results in paper, Table 1).
	
	Moreover, to train the baselines on the PoseTrack, which contains invisible joints, we perform a similar procedure we take for training our model which means if a joint disappears from ground-truth during training, no gradient for that joint is calculated. Besides, as the neck position is required for centering the pose for pose dynamics forecasting baselines, we tried our best to refine the dataset manually, to have a good estimation of neck in occluded cases and for other cases that the agent leaves the scene we completely discard the pose. During back propagation we simply ignore these samples (do not calculate loss for them) and in test time, we use the centered poses obtained from refined neck as input and the output is whatever model predicts. Important to note that we use the refined data only for centering the pose for input and the evaluation is performed with the original data.
	
	For the reported SC-MPF results in Table 1, we used the original SC-MPF code and metrics (requested from the authors). However, the PoseTrack data used in the SC-MPF paper is a very smaller subset of the dataset to ensure all joints for all persons in the selected sequences are fully visible as they did not model joint invisibility. We removed those assumptions from the input dataset, creating more realistic benchmarks, and used the whole dataset for the evaluation.

	\section*{E.~Experimental Settings}
	Regarding the objects used for H2O graph, we represent each object with four main features: 
	1) the extracted visual feature obtained from the detector
	2) together with its location defined as the center location of the extracted bounding box,
	3) the height and width of the bounding box, normalized over the sequence resolution and 
	4) the object class label as the final feature. The final object representations are obtained by passing these features through multiple MLP layers of sizes 5000, 1024 and 256. Similarly, The embedding dimensions of the MLP used for the context are 512 and 256.
	The hyper-parameters are selected through experiments on the validation set. We applied an initial learning rate of $5e^{-5}$ with a decay factor of $0.95$ and an Adam optimizer and the step size of 2 frames being injected in each step of curriculum learning to train the model. The cut off value ($\beta$) is set to be 200 pixels. 
	The GATs used are all single layer with 3 heads. 
	Each experiment is performed three times and their average values are reported. 
	
	\section*{F. Additional Results}
	Here we provide the results for an ablation study on the number of steps performed in the iterative message passing. Table \ref{tlb:msgablation} shows the results. As expected, when the number of message passing iterations increase the performance first improves and then starts declining. This is commonly explored by prior graph-based learning literature \cite{xu2019powerful}, a crucial aspect of the graph-level representation learning is that node representations become refined and more global with the increase of the number of iterations. Therefore, it is essential to find the sufficient number of iterations for the best performance, as outlined herein. 
	
	We also investigated the effect of using a sparse or dense graph as the input skeleton representation, which is connecting the human joints in compliance with the nature of human body skeleton or representing them as fully connected graphs and letting the model to learn their relationships.The results for this study is illustrated in Table \ref{tlb:inpablation}. The results indicate that the model can perform better when it learns the human joint relations by itself rather than sparse natural connections. This verifies the fact that the relationship between joints of an individual is not a simple hierarchical connection but every joint can have a segregated effect on each of the other joints directly.

	\begin{table}
		\footnotesize
		\begin{center}
			\caption{\footnotesize Error rate for ablation study on \textbf{3DPW} dataset (in cm) using different number of message passing iterations.}   \label{tlb:msgablation}
			\setlength{\tabcolsep}{4pt}
			\renewcommand{\arraystretch}{1.1}
			\begin{tabular}{c|c c c c c c }
				
				\hlineB{2}
				
				Message Passing& \multicolumn{5}{c}{milliseconds}\\
				
				\#iterations&100&240&500&640&900&AVG\\
				
				\hlineB{1.5}
				
				\textcolor{black}{1 iteration} &32.49&52.71&90.39&110.51&163.46&89.91\\
				
				\textcolor{black}{2 iterations} & 32.43&52.6&89.06&109.14&159.54&	88.55 \\
				
				\textcolor{black}{3 iterations} &\textbf{31.56}&\textbf{51.97}&\textbf{86.53}&\textbf{107.52}&\textbf{153.12}&\textbf{86.14}\\
				
				\textcolor{black}{4 iterations} &33.58&52.98&91.21&111.75&163.63&90.63\\

				\hlineB{2}

			\end{tabular}
		\end{center}
	\end{table}

	\begin{table}
		\footnotesize
		\begin{center}
			\caption{\footnotesize Error rate for ablation study on \textbf{3DPW} dataset (in cm) using a sparse or dense graph as input skeleton representation.}   \label{tlb:inpablation}
			\setlength{\tabcolsep}{4pt}
			\renewcommand{\arraystretch}{1.1}
			\begin{tabular}{c|c c c c c c }
				
				\hlineB{2}
				
				\multirow{2}{*}{Input representation} & \multicolumn{5}{c}{milliseconds}\\
				
				&100&240&500&640&900&AVG\\
				
				\hlineB{1.5}
				
				\textcolor{black}{Sparse Graph} &33.81&53.01&89.49&110.44&158.65&89.08\\
				
				\textcolor{black}{Dense Graph} &\textbf{31.56}&\textbf{51.97}&\textbf{86.53}&\textbf{107.52}&\textbf{153.12}&\textbf{86.14}\\

				\hlineB{2}

			\end{tabular}
		\end{center}
	\end{table}
	
\end{document}